\documentclass[10pt,twocolumn,letterpaper]{article}

\usepackage[pagenumbers]{cvpr} 

\usepackage{graphicx}
\usepackage{amsmath}
\usepackage{amssymb}
\usepackage[title]{appendix}
\usepackage{xspace}
\usepackage{mathtools}
\usepackage{bbding}
\usepackage{amsmath}
\usepackage{amsthm}
\usepackage{booktabs}
\usepackage{verbatim}
\usepackage{multirow}
\usepackage{makecell}
\usepackage{algorithm}
\usepackage{algorithmic}
\usepackage{times}
\usepackage{epsfig}
\usepackage{pifont}
\usepackage{booktabs}
\usepackage{capt-of}
\usepackage{colortbl}
\usepackage{dsfont}
\usepackage{comment}
\usepackage[dvipsnames]{xcolor}
\usepackage{pifont}
\usepackage{bbm}
\usepackage{url}
\usepackage{nicefrac}
\usepackage{colortbl}
\usepackage{cases}
\definecolor{cvprblue}{rgb}{0.21,0.49,0.74}
\usepackage[pagebackref,breaklinks,colorlinks,citecolor=cvprblue]{hyperref}

\usepackage[capitalize]{cleveref}
\crefname{section}{Sec.}{Secs.}
\Crefname{section}{Section}{Sections}
\Crefname{table}{Table}{Tables}
\crefname{table}{Tab.}{Tabs.}

\definecolor{baselinecolor}{gray}{.9}
\definecolor{darkgreen}{rgb}{0.13, 0.55, 0.13}

\renewcommand{\paragraph}[1]{\vspace{1.25mm}\noindent\textbf{#1}}

\let\originalleft\left
\let\originalright\right
\renewcommand{\left}{\mathopen{}\mathclose\bgroup\originalleft}
\renewcommand{\right}{\aftergroup\egroup\originalright}

\usepackage{caption}
\usepackage[switch]{lineno}

\def\paperID{1852} 
\def\confName{CVPR}
\def\confYear{2025}

\begin{document}


\newcommand{\method}{VideoAnydoor\xspace}

\title{\method: High-fidelity Video Object Insertion with Precise Motion Control}

\author{
    Yuanpeng Tu$^{1,2*}$ \quad
    Hao Luo$^{2,3}$ \quad
    Xi Chen$^{1}$ \quad
    Sihui Ji$^{1}$ \quad
    Xiang Bai$^{4}$ \quad
    Hengshuang Zhao$^{1,\dagger}$\\[2pt]
    $^{1}$The University of Hong Kong \quad
    $^{2}$DAMO Academy, Alibaba Group \quad
    $^{3}$Hupan Lab \quad
    $^{4}$HUST \\ [2pt]
    \textit{\href{https://videoanydoor.github.io/}{https://videoanydoor.github.io}}
}


\twocolumn[{
\renewcommand\twocolumn[1][]{#1}
\maketitle

\vspace{-33pt}
\begin{center}
    \centering
    \includegraphics[width=1.0\textwidth]{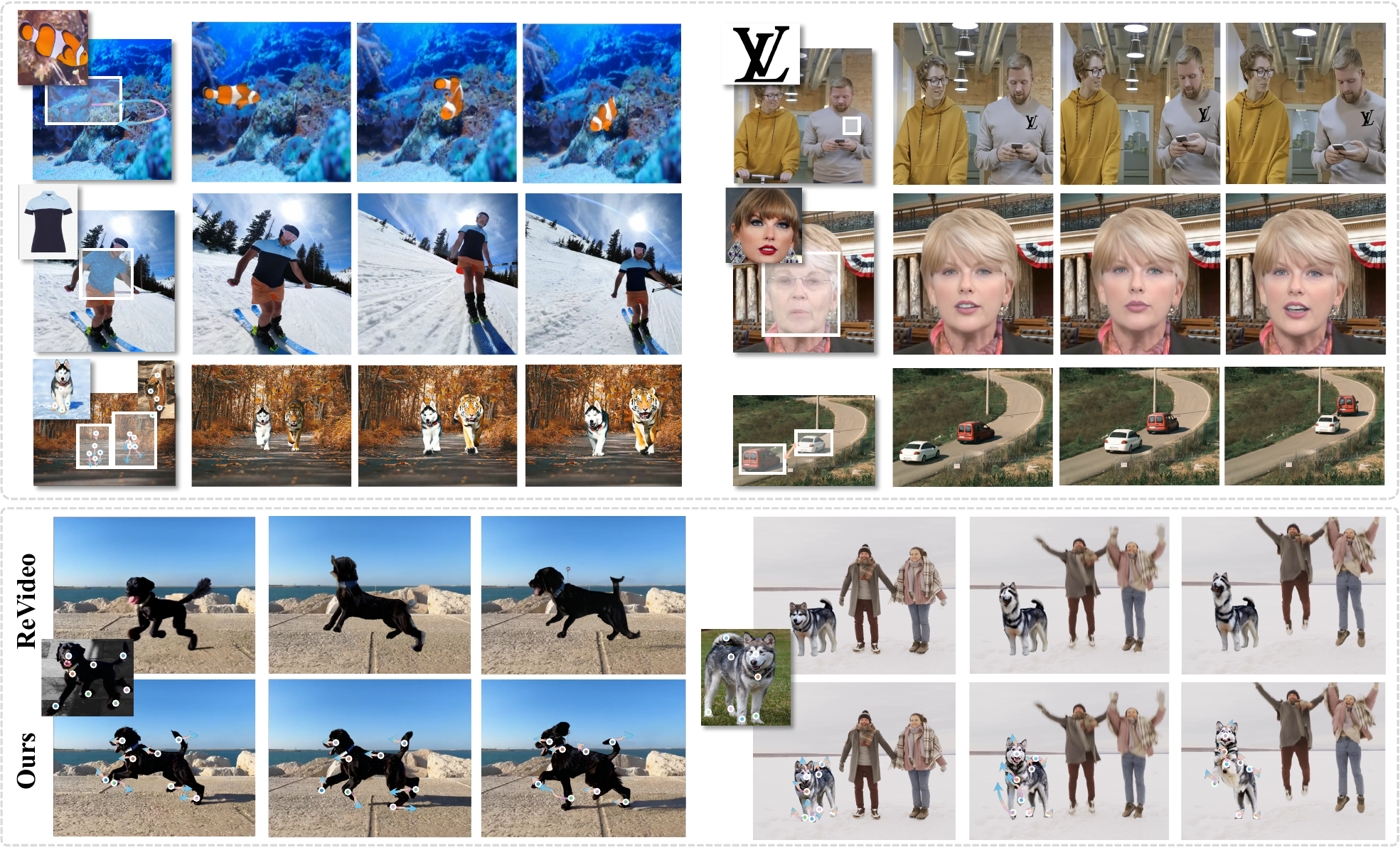}
    \captionof{figure}
    {
    \textbf{Demonstrations for video object insertion}. 
    \method preserves the fine-grained object details and enables users to control the motion with boxes or point trajectories. 
    Based on the robust insertion, users could further add multiple objects iteratively or swap objects in the same video. 
    Compared with the previous works, \method demonstrates significant superiority.} 
    \label{Fig:begin}
    \vspace{4pt}
\end{center}%
}]

\let\thefootnote\relax\footnotetext{*Work during DAMO Academy internship. $\dagger$ Corresponding author.}

\begin{abstract}
Despite significant advancements in video generation, inserting a given object into videos remains a challenging task.
%
%
The difficulty lies in preserving the appearance details of the reference object and accurately modeling coherent motion at the same time.
In this paper, we propose \method, a zero-shot video object insertion framework with high-fidelity detail preservation and precise motion control.  
Starting from a text-to-video model, we utilize an ID extractor to inject the global identity and leverage a box sequence to control the overall motion.
To preserve the detailed appearance and meanwhile support fine-grained motion control, we design a pixel warper. 
It takes the reference image with arbitrary key-points and the corresponding key-point trajectories as inputs.
It warps the pixel details according to the trajectories and fuses the warped features with the diffusion U-Net, thus improving detail preservation and supporting users in manipulating the motion trajectories.
In addition, we propose a training strategy involving both videos and static images with a weighted loss to enhance insertion quality.
\method demonstrates significant superiority over existing methods and naturally supports various downstream applications~(\textit{e.g.,} video face swapping, video virtual try-on, multi-region editing) without task-specific fine-tuning. 
\vspace{-5pt}
\end{abstract}

\vspace{-30pt}
\section{Introduction}
\label{sec:intro}
\vspace{-1mm}
The booming of diffusion models~\cite{SahariaCSLWDGLA22,kolors,li2023blip} has spurred significant advancements in text-to-video generation~\cite{sora,yang2024cogvideox, hong2022cogvideo} and editing~\cite{LiuZ00J24,TuDC0HWJ24,eva,JeongY24}. Some works~\cite{dreampose_2023,mimicmotion2024,wei2024dreamvideo,wang2024videocomposer} learn to edit the video based on posture~\cite{dreampose_2023,mimicmotion2024} or styles~\cite{liu2023stylecrafter} while other works~\cite{ku2024anyv2v,gu2024videoswap} explore modifying specific objects based on text descriptions.

In this paper, we investigate video object insertion, which means seamlessly placing a specific object~(with a reference image) into a given video with the desired motion and location. 
%
%
This ability has broad potential for real-world applications, like video composition, video virtual try-on, video face changing, \textit{etc}.

Although strongly in need, this topic remains under-explored by existing works. We analyze that the challenge of video object insertion mainly lies in two folds: accurate \textit{ID preservation} and precise \textit{motion control}. 
Recently, some works have made initial attempts in this field. 
AnyV2V~\cite{ku2024anyv2v} and ReVideo~\cite{mou2024revideo} leverage image composition model~\cite{chen2024anydoor} to insert the object in the first frame.   
Then, they propagate this modification to subsequent frames are under the guidance of text or trajectory control. 
However, this two-stage scheme may lead to suboptimal results if the first frame insertion is not satisfactory. 
Besides, as they do not inject ID information in the following frames, the object's identity and motion tend to collapse in the later frames.

Faced with this challenge, we attempt to accurately preserve the object's identity and precisely control the object's motion throughout the whole video. 
Specifically, we propose an end-to-end framework termed \textbf{\method}. 
Starting from a text-to-video diffusion model, the concatenation of random noise, object masks, and the masked video is utilized as input. Meanwhile, the reference image with no background is fed into the ID extractor to extract compact and discriminative ID tokens. Then these ID tokens are injected into the diffusion model together with the box sequence as coarse guidance of identity and motion to generate the desired composition. 
Additionally, a pixel warper module is designed for joint modeling of the fine-grained appearance and precise motion. It takes the reference image with arbitrary key-points and the corresponding key-point trajectories to warp the pixel details into the target regions according to the desired motion and pose. Moreover, to address the scarcity of high-quality videos, we manually augment extensive high-quality image data as videos to improve the fine-grained alignment of appearance details. As shown in Fig.~\ref{Fig:begin}, with these techniques, users can edit specific regions in the video by providing target images, drawing boxes and trajectory lines. It should be noted that our inserted object is not constrained by shape, appearance or the range of the given movement, demonstrating great robustness and generality in diverse scenarios.

Our contributions can be summarized as follows:
\begin{itemize}
\item We construct the first end-to-end video object insertion framework that supports both motion and content editing. 
Our framework seamlessly supports diverse applications, \textit{e.g.,} multi-region editing, video virtual try-on, and video face changing, \textit{etc.}

\item We propose pixel warper to warp the pixel details according to the desired motion. It takes the reference image with arbitrary key-points and the trajectories as inputs for fine-grained modeling of identity and motion, enabling accurate ID preservation and motion control. 

\item We design multiple strategies to further enhance the capability of accurate insertion, including image-video mix training, training trajectory filtering. Extensive experiments demonstrate their effectiveness in precise ID preservation and motion control.




\end{itemize}

\vspace{-2mm}
\section{Related Work}
\vspace{-2mm}
\paragraph{Image-level object insertion.} Generative object compositing~\cite{yang2022paint,ObjectStitch,yang2022paint,IMPRINT,ControlCom,chen2024anydoor} focus on implanting subjects in diverse contexts. Among these methods, Paint-by-Example~\cite{yang2022paint} proposes an information bottleneck to avoid the trivial solution. CustomNet~\cite{CustomNet} incorporates 3D novel view synthesis capabilities. IMPRINT~\cite{IMPRINT} decouples learning of identity preservation from that of compositing. AnyDoor~\cite{chen2024anydoor} utilizes a frequency-aware detail extractor to obtain detail maps. However, directly transferring similar insertion schemes as these to videos may result in imperfect performance as they fail to preserve fine-grained appearance details, while the quality of object insertion in videos is crucial for precise motion control. Nevertheless, these methods generally fail to insert objects with proper postures for motion control. Thus, to address these two issues, we conduct a detailed investigation on object insertion with accurate ID preservation and proper posture control.

\paragraph{Video editing.} Early methods~\cite{wu2023tune,ceylan2023pix2video,tokenflow2023,ouyang2024codef} primarily adopt training-free or one-shot tuning schemes owing to the lack of proper training data.
For example, Pix2Video~\cite{ceylan2023pix2video} first edits the first frame and then produces followed frames with cross-frame attention.
%
Recently, tuning-based methods~\cite{gu2024videoswap,wu2024customcrafter,ku2024anyv2v,mou2024revideo} have exhibited better results. Among them, text-prompt based schemes struggle to locate target regions. AnyV2V~\cite{ku2024anyv2v} uses an off-the-shelf image editing model to modify the first frame. Image-prompt based methods like ReVideo~\cite{mou2024revideo} design a three-stage training scheme that decouples content and motion control. VideoSwap~\cite{gu2024videoswap} uses semantic points to achieve video subject replacement. However, these methods either require extra fine-tuning, fail to keep the unedited region unchanged, or achieve poor motion/identity consistency with a two-stage scheme. To address these issues, we aim to design an end-to-end zero-shot video insertion framework that precisely modifies both content and motion according to user-provided instructions while keeping the unedited content unchanged in zero-shot.

\begin{figure*}[tbp]
    \centering
    \includegraphics[width=0.99\linewidth]{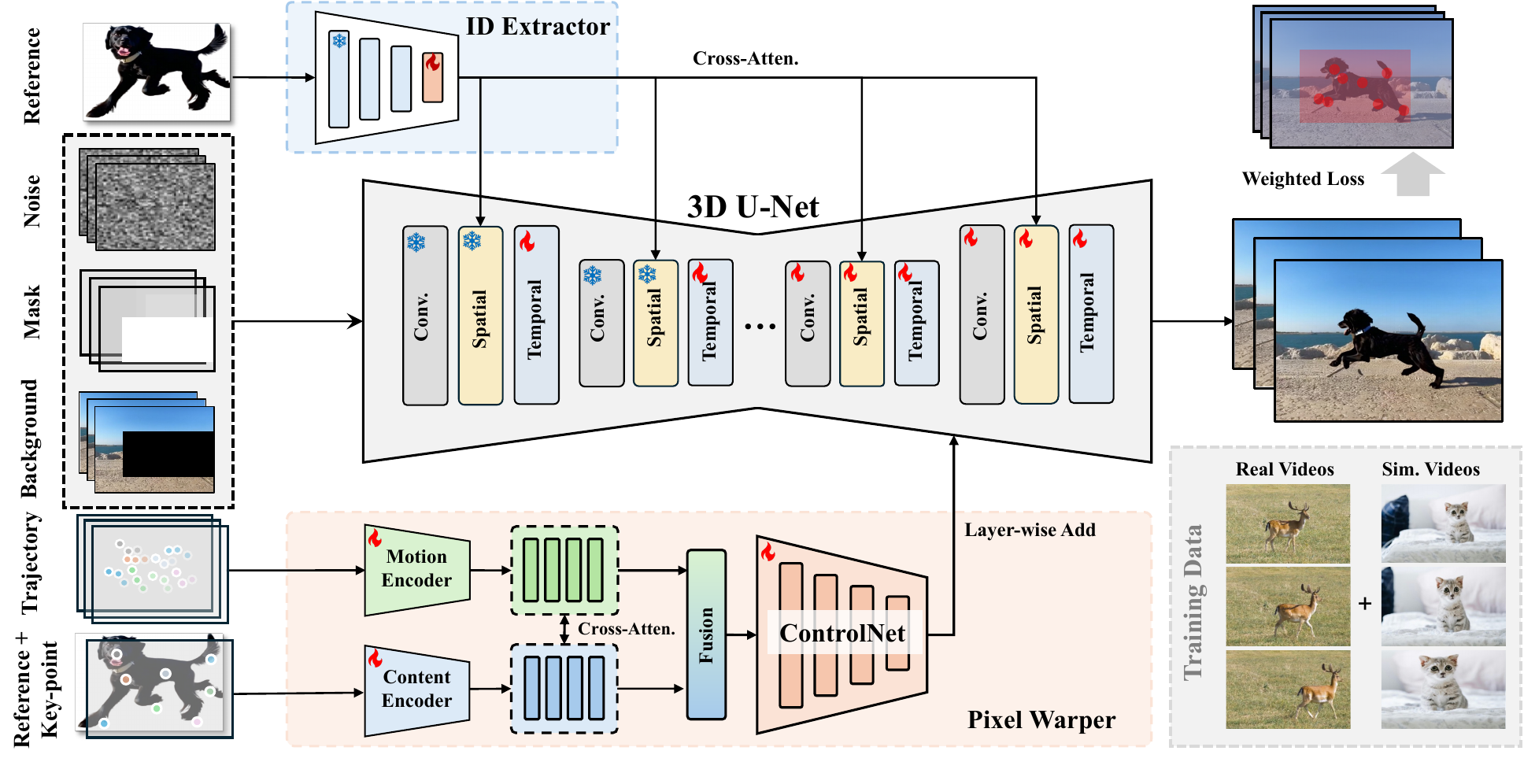}
    \vspace{-2mm}
    \caption{
    \textbf{The pipelines of our \method}. First, we input the concatenation of the original video, object masks, and masked video into the 3D U-Net. Meanwhile, the background-removed reference image is fed into the ID extractor, and the obtained features are injected into the 3D U-Net. In our pixel warper, the reference image marked with key points and the trajectories are utilized as inputs for the content and motion encoders. Then, the extracted embeddings are input into cross-attentions for further fusion. The fused results serve as the input of a ControlNet, which extracts multi-scale features for fine-grained injection of motion and identity. The framework is trained with weighted losses. We use a blend of real videos and image-simulated videos for training to compensate for the data scarcity.}  
    \label{Fig: framework}
\end{figure*}

\section{Method}
\vspace{-1mm}
\subsection{Overview of Framework}
\vspace{-1mm}
\paragraph{Task formulation.} In this paper, we focus on high-fidelity video object insertion, with the goal of subject insertion with user-provided trajectories, where the unedited regions should remain the same as the source video. The primary challenge of this task lies in aligning the motion trajectory of the given one while preserving the identity of the target concept, particularly its appearance details. 

\paragraph{Overall pipeline.} The \method pipeline is illustrated in Fig.~\ref{Fig: framework}. To reconstruct the background within the masked region, we build our method on a 2D in-painting diffusion model. Following the latent diffusion model~\cite{latent2023}, we encode both the source video and the masked video with a VAE encoder to obtain the latent space representations $z_{ori}$ and $z_{mask}$. The corresponding masks are 8 times down-sampled as the mask latent. Subsequently, DDIM inversion~\cite{DhariwalN21} is applied to transform the clean latent $z_{ori}$ back to the noisy latent $z_{T}$. Then we concatenate $z_{T}$, $z_{mask}$ and $z_{mask}$ as a 9-channel tensor for 3D U-Net. To utilize the priors of video generation, we integrate the motion layers~\cite{guo2023animatediff} into the in-painting model as the 3D U-Net to ensure essential temporal consistency. For coarse-grained control, we leverage the powerful visual encoder DINOv2~\cite{oquab2023dinov2} as the ID extractor for ID preservation and use the bounding boxes as motion guidance. Before feeding the reference image into the extractor, we remove its background with a segmentor~\cite{kirillov2023segany} and align the object to the image center to retain 
compact and ID-related representations. For fine-grained control, we adopt the interaction-friendly trajectory lines as the control signals and propose a pixel warper to warp the pixel details according to the desired motion for joint modeling of appearance details and precise motion. Finally, a weighted loss is used amplify the influence of key-points and design a novel image-video mix-training strategy to address the scarcity of high-quality video data.

For convenience, the trajectory map and correspondence reference image are denoted as $c_{mot} \in \mathcal{R}^{N\times 3\times H\times W}$ and $c_{ref-key} \in \mathcal{R}^{3\times H\times W}$ respectively. $V\in \mathcal{R}^{N\times 3\times H\times W}$, $A \in \mathcal{R}^{N\times 1\times H\times W}$ and $c_{ref} \in \mathcal{R}^{3\times H\times W}$ represent the original video, masks of the edited region and the reference image respectively. $N, H, W$ is the frame number, height, and width of the original video. The content and motion encoders are denoted as $E_c$ and $E_m$ respectively.

\paragraph{Inference configuration.} For users, they only need to provide a subject image, a source video, and a trajectory sequence. For the trajectory sequence, the users can directly use the trajectory of the object within the source video or just draw a start box and an end box to flexibly generate edited videos precisely aligned with the given conditions.

\vspace{-1mm}
\subsection{Pixel Warper}
\vspace{-1mm}
\begin{figure}[tbp]
    \centering
    \vspace{-1mm}
    
    \includegraphics[width=0.99\linewidth]{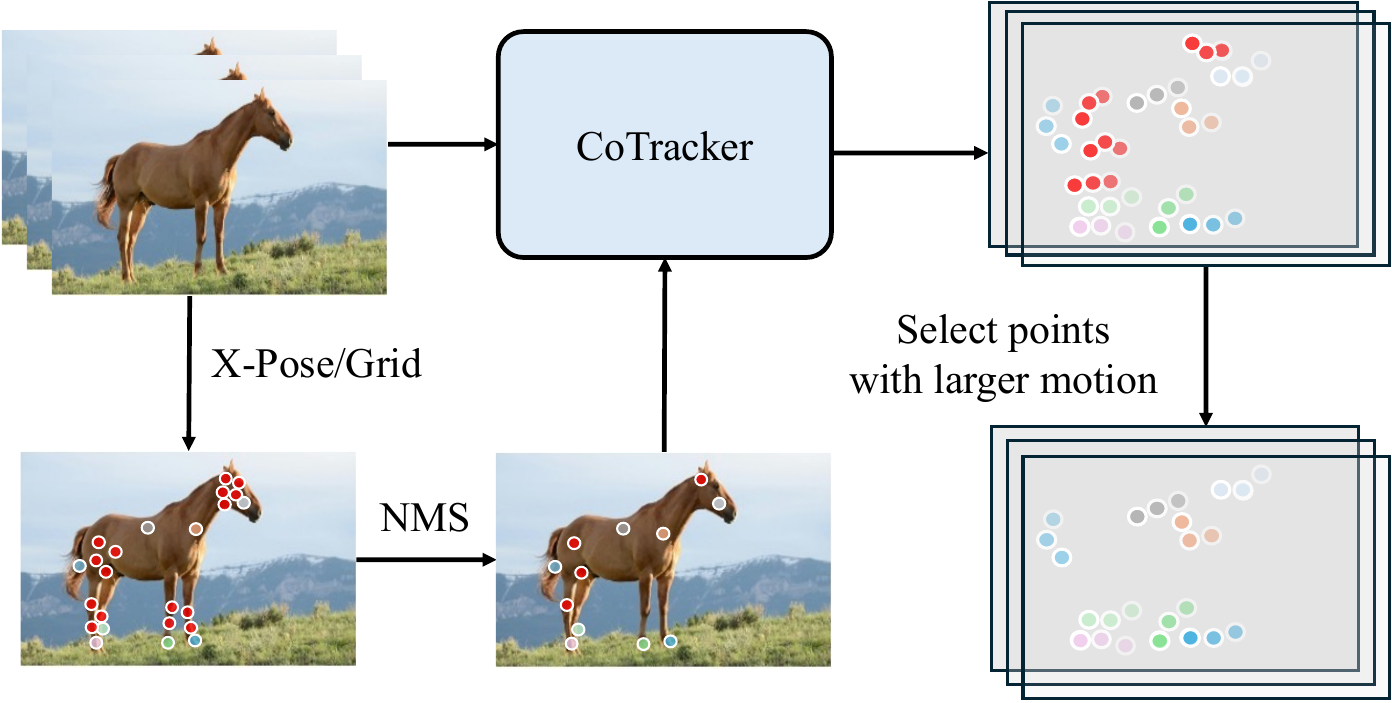}
    \vspace{-1mm}
    
    \caption{\textbf{Pipeline of trajectory generation for training data}. We first perform NMS to filter out densely-distributed points and then select points with larger motion. The retained ones can be sparsely distributed in each part of the target and contain more motion information, thus inducing more precise control.} \label{Fig: pointselect}
    \vspace{-2mm}
\end{figure}

\vspace{-1mm}
\paragraph{Trajectory sampling.} During training, it is essential to extract trajectories from videos to provide motion conditions. Previous works~\cite{xpose} show that the movement of objects can be controlled by general key-points. Thus, as shown in Fig.~\ref{Fig: pointselect}, we first input the first frame to X-Pose~\cite{xpose} to initialize the points for subsequent trajectory generation. For cases in which X-Pose fails to detect any key-points, we use a grid to sparsify dense sampling points. We empirically find that points with larger motion are more helpful for trajectory control. However, these points are mostly densely distributed in certain regions, resulting in severe information redundancy. Therefore, to filter out the undesired points, we first perform non-maximum suppression (NMS) to filter out points that are densely distributed. Then we apply motion tracking on each point to obtain their path lengths, \textit{e.g.}, $\{l_0,l_1,...,l_{N_{init}-1}\}$, where $N_{init}$ denotes the number of initial points. Then we retain $N$ points with the largest motion and use the corresponding trajectory map $c_{mot}$ as control signals. Different colors are assigned for $N$ points to represent different trajectories.

\paragraph{Motion injection.} A naive implementation of motion injection is only training a similar control module to inject the motion conditions as ~\cite{zhang2023adding, gu2024videoswap}. However, such a scheme may fail to accurately insert the objects with desired motion and appearance details, since it has no explicit semantic correspondence with the reference object. Thus the object may be inserted into the video with an undesired pose, leading to severe distortion in foreground regions. To address this issue, we input a pair of trajectory maps $c_{mot}$ and correspondence reference image $c_{ref-key}$ as fine-grained guidance. As show in Fig.~\ref{Fig: framework}, $c_{mot}$/$c_{ref-key}$ are first encoded by $E_c$/$E_m$ respectively. Then these two embeddings are input into two cross-attention modules respectively for semantic-aware fusion. Afterward, the fused two features are added and utilized as the input of a ControlNet~\cite{zhang2023adding} to extract multi-scale intermediate features $\{\mathrm{f}_c^{0}, \mathrm{f}_c^{0}, ..., \mathrm{f}_c^{P}\}$, where $P$ denotes the layer number of the diffusion model. Then these features are added to the corresponding layers for fine-grained modeling of appearance details and precise motion. This procedure can be formulated as:
\vspace{-2mm}
\begin{equation}
\mathbf{y}_c=\mathcal{F}\left(\mathbf{z}_t, t, \mathbf{c}_{ref}; \Theta\right)+\mathcal{Z}\left(\mathcal{F}\left(\mathbf{z}_t+\mathcal{Z}\left(\mathbf{f}_c\right), t, \mathbf{c}_{ref} ; \Theta_c\right)\right),
\end{equation}
where $\mathbf{y}_c$ denotes the new diffusion features. $\mathcal{Z}$ represents the function of zero-conv~\cite{ZhangRA23}. $\Theta_c$ and $\Theta$ are the parameters of the ControlNet and the diffusion model.  

\paragraph{Weighted loss.} Inspired by ~\cite{wei2024DreamVideo2}, we try to adopt a reweight scheme to improve the fine-grained modeling of identity and motion. Specifically, we solely perform reweighting on the region surrounding the trajectory. For trajectories with relatively large motion amplitudes, we apply greater weights to achieve more accurate motion control. Denote the ratio of the region that $i$-th trajectory covers as $R_i$, we perform 8 times down-sampling on the region, which is denoted as $\mathbf{A_{trj}^i}$. The weighted loss can be formulated as:

\vspace{-4mm}
\begin{equation}
\begin{split}
\mathcal{L}= \sum_{i = 1}^{N} (
({\lambda R_i\mathbf{A_{trj}^i}+{(1-\mathbf{A_{trj}^i})/N}})\cdot\left\|\delta- \delta^*\right\|_2^2),
\end{split}
\end{equation}
\vspace{-2mm}

\noindent where $\lambda$ denotes the balancing loss weight. $\delta^{*}, \delta$ are the prediction of the 3D U-Net and the target.

\vspace{-1mm}
\subsection{Training Strategy}
\vspace{-1mm}

\begin{table}[t]
\caption{%
    \textbf{Statistics of datasets} used for training our \method. ``quality'' particularly refers to the image resolution.
}
\label{tab:datasets}   
\centering\footnotesize
\resizebox{1.0\linewidth}{!}{
\setlength{\tabcolsep}{3pt}
\begin{tabular}{lcccc}
\toprule
  Dataset                                   & Type  & \# Samples  & Mask Quality  & Video Quality  \\
\midrule
YouTubeVOS~\cite{Yang2019vis}&     Video          &    4,453          & High    & Low  \\
YouTubeVIS~\cite{Yang2019vis}   &         Video      &      2,883            & High    & Low  \\
UVO~\cite{du20211st}       &          Video     &       10,337           & High    & Low  \\
MOSE~\cite{MOSE}    &          Video      &        2,149          &  High   & High  \\
VIPSeg~\cite{miao2022large}                        & Video              &   3,110               &High     &High   \\
VSPW~\cite{miao2021vspw}                 &       Video        &        3,536          & High    &  High \\
SAM2~\cite{ravi2024sam2}                 &       Video        &        51,000          & High    &  High \\
Pexel                       &  Video             & 6,000                  &Medium     & High  \\
MVImgNet~\cite{yu2023mvimgnet}                        & Video              &  219,188                & High    & High  \\
ViViD~\cite{fang2024vivid}                        &  Video             & 9,700                 & High    & High  \\
CHDTF~\cite{zhang2021flow}                        &   Video            &      362            &High     & High  \\
CelebV-HQ~\cite{zhu2022celebvhq}                        &   Video            &      35,666             &High     & High  \\
Pexel                        &   Image            &   95,000               & Medium    & High  \\
\bottomrule
\end{tabular}}
\vspace{-2mm}
\end{table}


\begin{figure*}[tbp]
    \centering
    \includegraphics[width=0.99\linewidth]{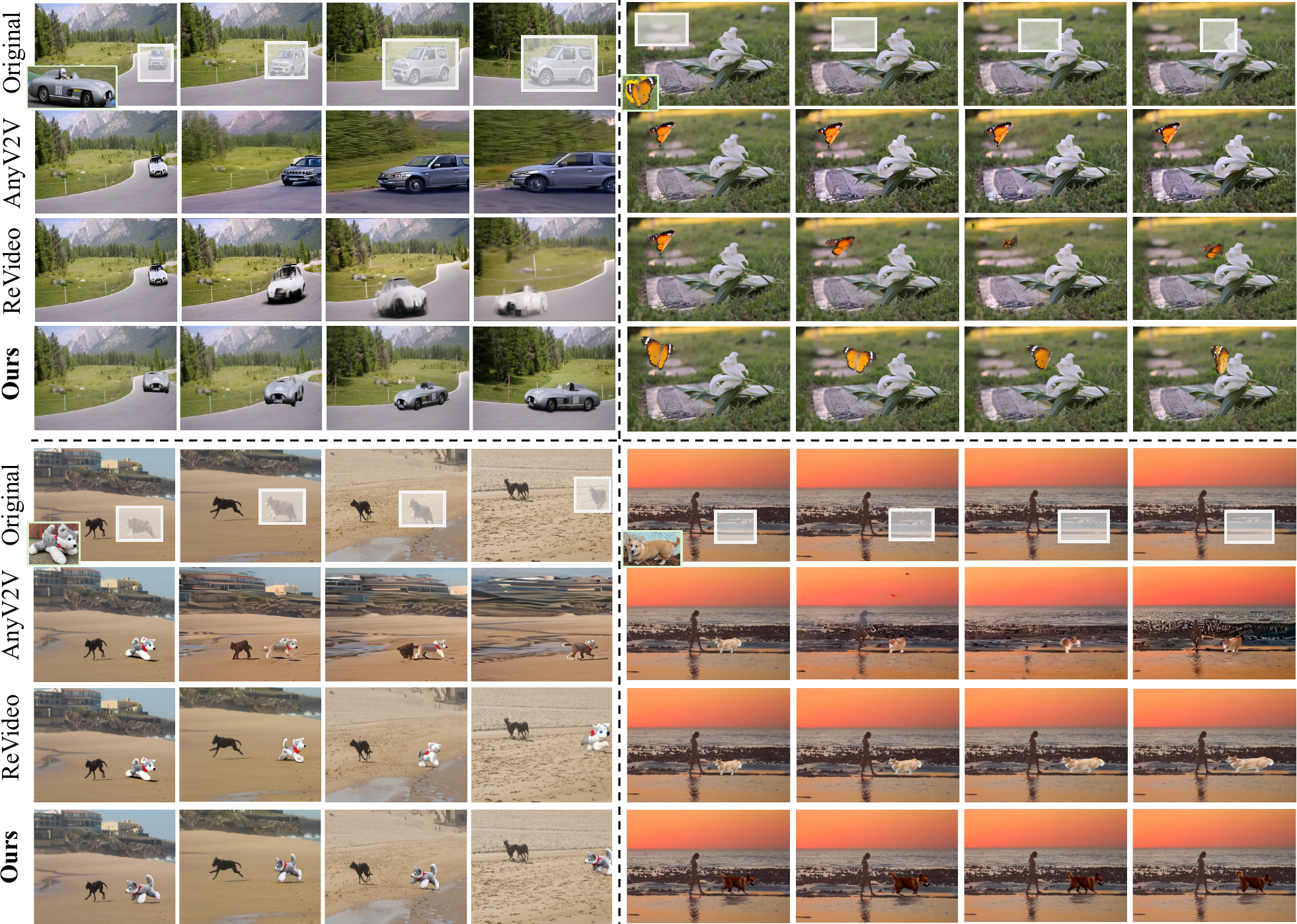}
    \caption{\textbf{Comparison results} between \method and existing state-of-the-art video editing works. Our \method can achieve superior performance on precise control of both motion and content.} \label{Fig: comparison}
    \vspace{-2mm}
\end{figure*}

\vspace{-1mm}
\paragraph{Dataset preparation.} The ideal training samples are video pairs for ``different objects in the same scene'', which are hard to collect with existing datasets. As alternatives, we sample all the needed data from the same video. Specifically, for a video, we pick a video clip and a frame that has the largest distance from the clip, which is assumed to contain the most dissimilar object from the one in the video clip. We take their masks for the foreground objects and remove the background for the select frame. Then we crop it around the mask as the target object. For the video clip, we generate the box sequence and remove the box region to get the scene video, where the unmasked video could be used as the training ground truth. Specifically, we use the expanded bounding box rather than the tightly-surrounded one in implementation. For boxes with a small moving range, we use the union of the boxes as the final box to reduce the impact of the bounding box on the motion. The full data for training is shown in Tab.~\ref{tab:datasets}, which covers both videos from diverse domains and high-quality images to compensate for the scarcity of high-quality videos.

\vspace{-1mm}
\paragraph{Image-video mixed training.} Different from previous works~\cite{chen2024videocrafter2} that utilize high-quality images for two-stage disentangled training, we resort to employing them for joint training with videos. However, directly repeating images will impair the discriminative learning of temporal modules. Instead, we augment the images as videos with manual camera operation. Specifically, we either randomly translate the image at equal intervals from different directions, or gradually crop the original image at equal intervals to obtain an image sequence. Then the image sequence is processed with bilinear interpolation to enhance the video smoothness. Although the augmented videos would benefit the learning of appearance variation, the essential difference between them and real videos will potentially impair motion learning. Thus similar to~\cite{chen2024anydoor}, we adopt adaptive timestep sampling to enable different modalities of data to contribute to different stages of denoising training.

\begin{figure*}[tbp]
    \centering
    \includegraphics[width=1.0\linewidth]{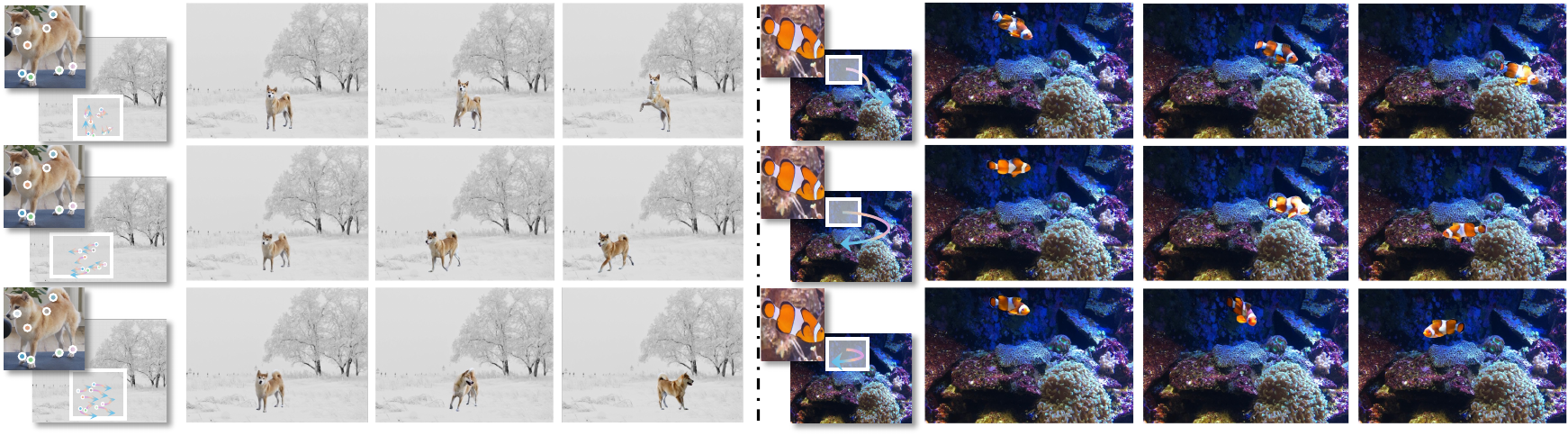}
    \caption{\textbf{Demonstrations for precise motion control}. \method can achieve precise alignment with the given trajectories and objects when using a pair of reference images marked with key-points and corresponding trajectory maps as input.} 
    \label{Fig: comparison_action}
\end{figure*}

\vspace{-2mm}

\section{Experiments}

\subsection{Experimental Setup}

\paragraph{Implementation details.} In this work, we choose Stable Diffusion XL with motion modules as the base generator. During training, we process the image resolution to 512$\times$512 and adopt Adam optimizer with an initial learning rate of $1e^{-5}$. We use DDIM for 50-step sampling and classifier-free guidance with a cfg of 10.0 for inference. The model is optimized for 120K iterations on 16 NVIDIA A100 GPUs with a batch size of 32. We only use 8 points for trajectory generation of each sample. In actual use, these parameters can be adjusted by the user according to different subjects and the desired generation effect. 

\paragraph{Benchmarks.} For comprehensive evaluation, we construct a benchmark consisting of around 200 videos collected from Pexel, which includes ten different categories (\textit{e.g.,} persons, dogs). We also make qualitative analysis on the ViViD~\cite{fang2024vivid} and CHDTF~\cite{zhang2021flow} test set to evaluate the performance for virtual video try-on and talking head generation.

\paragraph{Evaluation metrics.} On our constructed benchmarks, for evaluation of ID preservation, we calculate both CLIP-Score~\cite{Radford2021LearningTV} and DINO-Score~\cite{oquab2023dinov2} to reflect the similarity between the edited region and the target subject, where PSNR~\cite{PSNR} is employed to measure the reconstruction quality of unedited regions as well. Additionally, for customized concepts, we follow Custom Diffusion~\cite{kumari2022customdiffusion} to compute pairwise image alignment between each edited frame and each reference concept image. In addition, we feed the edited videos to Cotracker~\cite{karaev2023cotracker} to calculate the tracking metrics~\cite{karaev2023cotracker} with the points in original videos as ground truth labels. Finally, we organize user studies with a group of 15 annotators to rate the edited results from the perspective of quality, fidelity, fluidity of movement, and diversity.

\begin{figure}[tbp]
    \centering
    \includegraphics[width=1.0\linewidth]{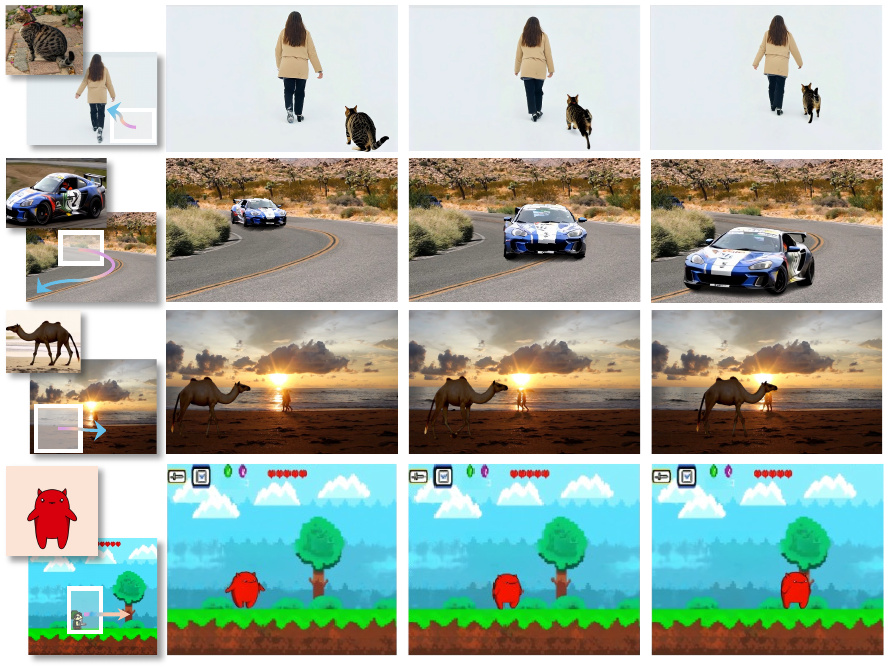}
    \vspace{-15pt}
    \caption{\textbf{More visual examples} of \method. It preserves fine-grained details (\textit{e.g.}, logos on the car) and achieves smooth motion control (\textit{e.g.}, the tail of the cat) with our pixel warper.} \label{Fig: comparison_moreexample}
    \vspace{-2mm}
\end{figure}

\begin{table}[t]
\caption{%
    \textbf{Quantitative comparison} between our \method and other related work. Six automatic metrics are employed for the performance evaluation of both content and motion. \method outperforms these methods across all the metrics.
}
\label{tab:automatic_psnr}
\centering
\resizebox{1.0\linewidth}{!}{
\setlength{\tabcolsep}{1pt}
\begin{tabular}{lcccccc}
\toprule
                                     & PSNR~($\uparrow$)   & CLIP-Score~($\uparrow$)  & DINO-Score~($\uparrow$) & AJ~($\uparrow$)   & ${\delta}_{avg}^{vis}$~($\uparrow$)  & OA~($\uparrow$)\\
\midrule
ConsistI2V~\cite{ren2024consisti2v}           &25.1                 &  64.7                &   40.6    & 49.3               &51.1              &57.2   \\
I2VAdapter~\cite{guo2023i2v}           &  24.3               & 67.1                 & 42.2     & 51.2               &53.7              &59.9    \\
AnyV2V~\cite{ku2024anyv2v}     &     30.1            &  70.2               &   47.2   & 54.1               &55.8              &61.1   \\
ReVideo~\cite{mou2024revideo}       & 33.5                & 74.2                 &51.7       & 79.2               &81.4              &83.2  \\
\method (ours)                  & \textbf{38.0}                &\textbf{81.4}                  & \textbf{59.1}   &\textbf{88.3}               &\textbf{91.5}              &\textbf{92.5}    \\
\bottomrule
\end{tabular}}
\end{table}

\subsection{Qualitative Comparison}
Among existing methods, ReVideo~\cite{mou2024revideo} heavily relies on the first frame modified by existing image customization methods and it adopts a semantic-unaware motion injection manner. AnyV2V~\cite{ku2024anyv2v} adopts a similar two-stage generation scheme with text prompts. In Fig.~\ref{Fig: comparison}, we can observe that AnyV2V~\cite{ku2024anyv2v} suffers from content distortions for both the edited and unedited areas. Moreover, it has poor motion consistency due to using texts as control signals. For ReVideo, there exists an obvious loss of edited content as well, especially for the cases with large motion. It exhibits inferior pose control over the inserted object owing to the lack of semantic information within motion signals. In comparison, our \method can effectively preserve the unedited content while allowing users to customize the motion in editing areas. We provide more examples in Fig.~\ref{Fig: comparison_action} and Fig.~\ref{Fig: comparison_moreexample}, where we insert the same object with different trajectories and different objects in diverse scenarios.

\begin{figure*}[tbp]
    \centering
    \includegraphics[width=1.0\linewidth]{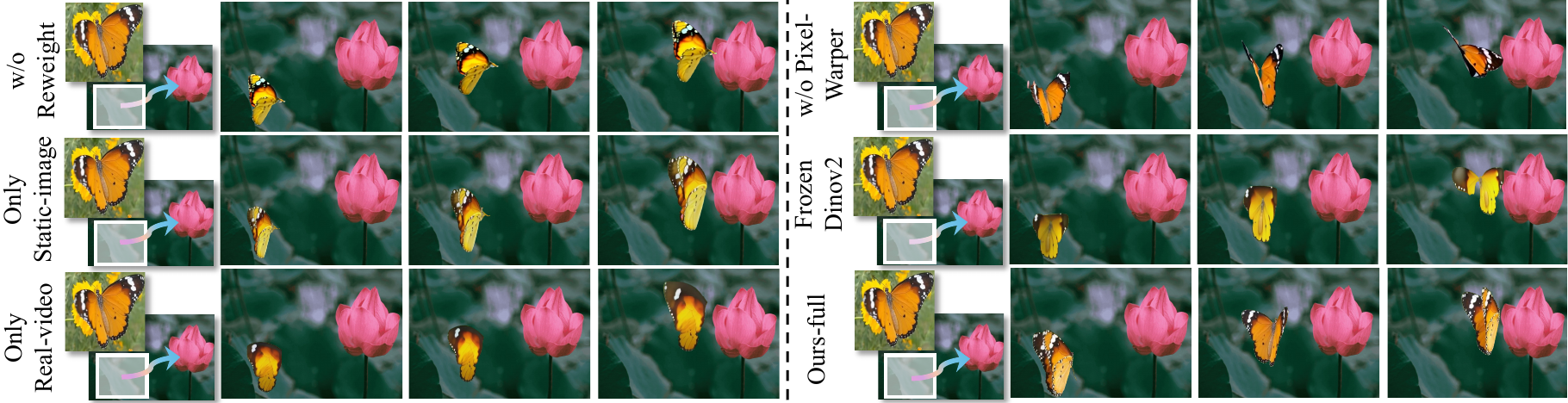}
    \caption{\textbf{Qualitative ablation studies} on the core components of \method. When removing the pixel warper, it suffers from poor motion consistency due to the undesired posture. And it can be observed that all the components contribute to the best performance.} \label{Fig: comparison_ablation}
\end{figure*}
  
\subsection{Quantitative Comparison}
\paragraph{ID preservation.} We first conduct quantitative evaluation with CLIP-Score~\cite{Radford2021LearningTV}, DINO-Score~\cite{oquab2023dinov2}, PSNR~\cite{PSNR}. Previous approaches impose heavy reliance on the existing image customization methods to acquire the first frame, making them retain the distortions within the first frame for subsequent generation. Thus it can be observed in Tab.~\ref{tab:automatic_psnr} that they generally achieve inferior results to our method. Moreover, since there is no explicit condition for AnyV2V~\cite{ku2024anyv2v}, ConsistI2V~\cite{ren2024consisti2v}, I2VAdapter~\cite{guo2023i2v} to keep the unedited regions unchanged, these methods perform much worse than our method and ReVideo on PSNR. Overall, our method achieves a clear advantage over the compared methods across all the metrics.

\paragraph{Motion consistency.} We conduct further quantitative experiments with the metric in tracking tasks to evaluate the motion alignment. Specifically, we track the key-points in the first frame of the original video in the edited one with the Cotracker model~\cite{karaev2023cotracker} and adopt the original trajectories as ground truths. The results are summarized in Tab.~\ref{tab:automatic_psnr}. Due to the lack of explicit motion control, AnyV2V, I2VAdapter, and ConsistI2V usually generate static or distorted motion for the edited content. Compared with them, \method demonstrates the best performance. Besides these results, we provide evaluation from aesthetic and technical views in the Appendix as well.

\begin{table}[t]
\caption{%
    \textbf{User study} on the comparison between our \method and existing alternatives.
    ``Quality'', ``Fidelity'', ``Smooth'', and ``Diversity'' measure synthesis quality, object identity preservation, motion consistency, and object local variation, respectively.
    Each metric is rated from 1 (worst) to 4 (best).
}
\label{tab:userstudy}
\centering\footnotesize
\setlength{\tabcolsep}{1.5pt}
\begin{tabular}{lcccc}
\toprule
                                        & Quality~($\uparrow$) & Fidelity~($\uparrow$) & Smooth~($\uparrow$) & Diversity~($\uparrow$)  \\
\midrule
ConsistI2V~\cite{gu2024videoswap}        & 1.80               & 1.75              &  2.30 &1.50\\
AnyV2V~\cite{ku2024anyv2v}                   &1.90           &1.85  &1.50 & 2.10 \\
ReVideo~\cite{mou2024revideo}              &2.65      &2.55       & 2.50 &2.25 \\
\method (ours)                          & \textbf{3.75}     &\textbf{3.80}     & \textbf{3.65} & \textbf{3.70}\\
\bottomrule
\end{tabular}
\end{table}

\begin{table}[t]
\caption{%
    \textbf{Quantitative evaluation} of core components in \method on ID preservation. $\dagger$ denotes removing the semantic points in the key-point image. 
}
\label{tab:automatic_mix_id}
\centering
\resizebox{1.0\linewidth}{!}{
\setlength{\tabcolsep}{1pt}
\begin{tabular}{lccc}
\toprule
                 & PSNR~($\uparrow$)   & CLIP-Score~($\uparrow$)  & DINO-Score~($\uparrow$)  \\
\midrule
Only Real-video Data          & 34.6                 &  75.0                & 51.7   \\
Only Static-image Data         & 33.9                 &  73.8                & 51.2   \\
\hline

FrozenDINOv2        & 33.5               & 74.3                 &51.6    \\

w/o $\rm Pixel Warper^{\dagger}$            &  35.3              & 77.4                 & 53.0   \\
w/o Pixel Warper           &  33.8              & 72.4                 & 48.5   \\
w/o Weighted Loss           &  35.1              & 77.0                 & 53.1   \\
w/ Box Loss~\cite{wei2024DreamVideo2}           &  37.7              & 81.1                 & 58.9   \\
\hline
Ours-full          &  \textbf{38.0}               & \textbf{81.4}                 & \textbf{59.1}   \\
\bottomrule
\end{tabular}}
\end{table}


\paragraph{User study.} We organize a user study to compare ConsistI2V~\cite{ren2024consisti2v}, ReVideo~\cite{mou2024revideo}, AnyV2V~\cite{ku2024anyv2v}, and our \method. Specifically, we let 20 annotators rate 20 groups of videos, where each group contains the original video and four edited videos. For each group, we provide one image edited by AnyDoor~\cite{chen2024anydoor} as the first frame for the compared methods. Besides, we provide detailed regulations to rate the generated videos for scores of 1 to 4 from four views: ``Quality'', ``Smooth'', ``Fidelity'', ``Diversity''. ``Fidelity'' measures ID preservation, and ``Quality'' counts for whether the result is harmonized without considering fidelity. ``Smooth'' assesses the motion consistency. We use ``Diversity'' to measure the differences among the synthesized results. The user-study results are shown in Tab.~\ref{tab:userstudy}. It can be noted that our model demonstrates significant superiority, especially for ``Fidelity'', and ``Smooth''. Such results fully verify the effectiveness of our method.

\begin{figure*}[tbp]
    \centering
    \includegraphics[width=1.0\linewidth]{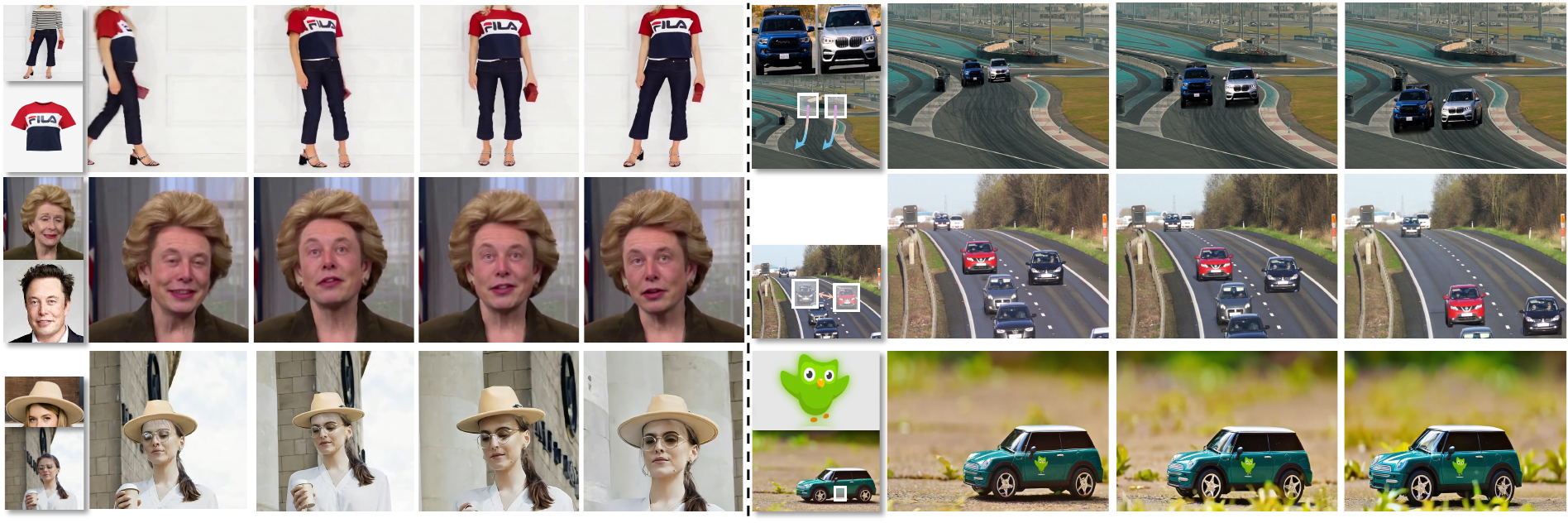}
    \vspace{-2mm}
    
    \caption{\textbf{More applications of \method}. Our framework seamlessly supports various tasks like video virtual try-on, talk head generation, multi-region editing, \textit{etc.} The results show that \method could effectively preserve the structure and identity and impose precise control on movements of multiple objects in diverse scenarios.} \label{Fig: talkhead}
    \vspace{-2mm}
\end{figure*}

\subsection{Ablation Studies}


\begin{table}[t]
\caption{%
    \textbf{Quantitative evaluation}  of core components in \method on motion consistency. $\dagger$ denotes removing the semantic points in the key-point image. 
}
\label{tab:automatic_mix_motion}
\centering
\resizebox{1.0\linewidth}{!}{
\setlength{\tabcolsep}{9pt}
\begin{tabular}{lccc}
\toprule
                 & AJ~($\uparrow$)   & ${\delta}_{avg}^{vis}$~($\uparrow$)  & OA~($\uparrow$)\\
\midrule
Only Real-video Data         &66.5  &67.5 & 69.9\\
Only Static-image Data          & 71.4  &72.0   &74.3   \\ \hline
$ \rm Frozen  DINOv2$   & 80.0               &82.4             &85.3  \\
w/o $\rm Pixel Warper^{\dagger}$           & 80.3               &81.3             &84.4  \\
w/o $\rm Pixel Warper$           & 78.5               &81.7            &83.7  \\
w/o Weighted Loss           &  75.4              & 84.2                 & 85.1   \\
w/ Box Loss~\cite{wei2024DreamVideo2}  &  88.0              & 91.1                 & 92.3   \\
\hline
Ours-full         & \textbf{88.3}              &\textbf{91.5}              &\textbf{92.5}  \\
\bottomrule
\end{tabular}}
\end{table}

\paragraph{ID preservation.} We conduct an investigation of the core components on ID preservation. From Tab.~\ref{tab:automatic_mix_id}, we can observe that training with fixed DINOv2 induces much inferior performance. Training only on videos suffers from severe accuracy degradation across all the metrics. Moreover, our pixel warper can effectively help inject the appearance details according to the motion. The results also show that the weighted loss is superior to the box loss~\cite{wei2024DreamVideo2} and can bring a performance boost to the baseline by making the model focus on the foreground regions. The best performance can be achieved by training with all the modules.

\paragraph{Motion consistency.} We present evaluation outcomes of the motion control for core components in Tab.~\ref{tab:automatic_mix_motion}. Results show that training with static images causes a significant accuracy drop due to impaired temporal module learning and inferior motion consistency when only using key-point trajectories in the pixel warper. Training with image-simulated videos aids precise motion control, likely due to facilitating fine-grained appearance reconstruction and making inter-frame key-point correspondence easier. Beyond this, we conduct more comparisons with different variants of pixel warper in Tab.~\ref{tab:automatic_mix_pw_detail}. It can be observed that selecting points with larger motion gives a significant performance boost. Moreover, using all grid-sampled points is inferior to extracted key-points. Selecting samples without considering distance leads to inferior performance as points are densely distributed in certain regions as well. Training with loosely-surrounded boxes leads to precise motion control for key-point trajectories. Qualitative comparisons in Fig.~\ref{Fig: comparison_ablation} show a similar phenomenon to quantitative results. All modules contribute to the best performance.

\begin{table}[t]
\caption{%
    \textbf{Detailed quantitative evaluation}  of the pixel warper in \method on motion consistency.  ``Tight box'' denotes training with tightly-surrounded boxes. 
}
\label{tab:automatic_mix_pw_detail}
\centering
\resizebox{1.0\linewidth}{!}{
\setlength{\tabcolsep}{10pt}
\begin{tabular}{lccc}
\toprule
                 & AJ~($\uparrow$)   & ${\delta}_{avg}^{vis}$~($\uparrow$)  & OA~($\uparrow$)\\
\midrule
Random X-Pose points & 80.6  &82.4 &83.0 \\
Grid points & 82.4 & 83.3& 85.0\\
w/o NMS &82.2  &83.0 &84.3 \\
Tight box & 83.4 & 85.6& 86.3\\
\hline
Ours-full            & \textbf{88.3}              &\textbf{91.5}              &\textbf{92.5}  \\
\bottomrule
\end{tabular}}
    \vspace{-2mm}

\end{table}

 \vspace{-1mm}
\subsection{More Applications}
    \vspace{-1mm}

\noindent\textbf{Virtual video try-on.} As shown in Fig.~\ref{Fig: talkhead}, without extra task-specific tuning, our \method demonstrates satisfactory performance for virtual try-on on the ViViD dataset~\cite{fang2024vivid}, where diverse patterns of the target clothes can be well preserved across different frames. Such results underscore the strong generalization abilities of our method.

\noindent\textbf{Talking head generation.} Besides video try-on, we further apply our method to talking head generation, which requires more precise control of tiny movements and preservation of face identities. Specifically, we conduct evaluations on the CHDTF dataset~\cite{zhang2021flow} and use 16 points with the largest movement in the face landmarks as the initial 
key-points. As demonstrated in Fig.~\ref{Fig: talkhead}, \method could give satisfactory performance for this task as well.

\noindent\textbf{Multi-region editing.} In addition, we extend our \method to multi-region editing. As shown in Fig.~\ref{Fig: talkhead}, we can achieve precise control of multi-object insertion for both motion and content. Besides, it can be used for object swapping and inserting logos or ornaments as well. As shown in Fig.~\ref{Fig: talkhead}, we can precisely place the hat on the head and achieve smooth posture control.

\section{Conclusion}
In this paper, we present \method for end-to-end video object insertion with precise motion control. Specifically, it can effectively characterize the reference target with an ID extractor when trained with a combination of videos and high-quality images. Moreover, it can achieve smooth motion consistency and effective preservation of appearance details through the proposed pixel warper. Our \method has promising performance on both video object insertion task and diverse precise video editing applications, \textit{e.g.}, (1) object insertion, (2) virtual video try-on, (3) video face swapping, and (4) multi-region editing. Extensive qualitative and quantitative experimental results demonstrate its superiority over previous methods on the alignment of both motion and identity with the given control signals. It provides a universal solution for general region-to-region mapping tasks as well.

\paragraph{Limitations.} Despite impressive results, our method still struggles with complex logos. This issue might be solved by collecting related data or using stronger backbones.

\paragraph{Acknowledgements.} This work was supported by DAMO Academy via
DAMO Academy Research Intern Program.


\medskip

{\small
\bibliographystyle{ieee_fullname}
\bibliography{egbib}
}

\end{document}